\documentclass[letterpaper, 10 pt, conference]{ieeeconf}  

\IEEEoverridecommandlockouts                              

\usepackage{times}

\usepackage{multicol}
\usepackage[bookmarks=true]{hyperref}
\usepackage{amsmath,amssymb,amsfonts}
\usepackage{booktabs}
\usepackage{graphicx}
\usepackage{xcolor}
\usepackage{bbm}
\usepackage{nicefrac}
\usepackage{comment}
\usepackage[font=footnotesize]{caption}
\usepackage{algorithm}
\usepackage{algpseudocode}
\usepackage{cite}

\newtheorem{definition}{Definition}

\newcommand{\sref}[1]{Sec. \ref{#1}}
\newcommand{\figref}[1]{Fig. \ref{#1}}

\def\AA{\mathcal{A}}
\def\RR{\mathbb{R}}

\def\TT{\mathcal{T}}
\def\DD{\mathcal{D}}
\def\SS{\mathcal{S}}
\def\MM{\mathcal{M}}
\def\ZZ{\mathcal{Z}}

\def\SC{\mathcal{SC}}
\def\LL{\mathcal{L}}

\title{\LARGE \bf
Multi-Agent Strategy Explanations for Human-Robot Collaboration
}

\author{Ravi Pandya$^*$, Michelle Zhao$^*$, Changliu Liu, Reid Simmons, Henny Admoni
\thanks{*contributed equally to this work.}
\thanks{Authors are with the Robotics Institute at Carnegie Mellon University, Pittsburgh, Pennsylvania, \tt\small \{rapandya, mzhao2, cliu6, rsimmons, hadmoni\}@andrew.cmu.edu}
}

\begin{document}

\maketitle
\thispagestyle{plain}
\pagestyle{plain}

\begin{abstract}
As robots are deployed in human spaces, it is important that they are able to coordinate their actions with the people around them. Part of such coordination involves ensuring that  people have a good understanding of how a robot will act in the environment. This can be achieved through explanations of the robot's policy. Much prior work in explainable AI and RL focuses on generating explanations for single-agent policies,
but little has been explored in generating explanations for \textit{collaborative} policies. In this work, we investigate how to  generate multi-agent strategy explanations for human-robot collaboration. We formulate the problem using a generic multi-agent planner, show how to generate visual explanations through strategy-conditioned landmark states and generate textual explanations by giving the landmarks to an LLM. 
Through a user study, we find that when presented with explanations from our proposed framework, users are able to better explore the full space of strategies and collaborate more efficiently with new robot partners.
\end{abstract}

\IEEEpeerreviewmaketitle

\section{Introduction}
Imagine you have just packed up your belongings to move into a new house. You and a friend have to move some large boxes out to the moving van, but it's impossible to hold boxes and open doors along the way yourself. How would you decide the right strategy for the two of you to efficiently move all the boxes outside? You will likely briefly stop to discuss the different ways you might coordinate to hold the doors and move the boxes out, then settle on a plan before moving everything.

\begin{figure}
    \centering
    \includegraphics[width=0.95\columnwidth]{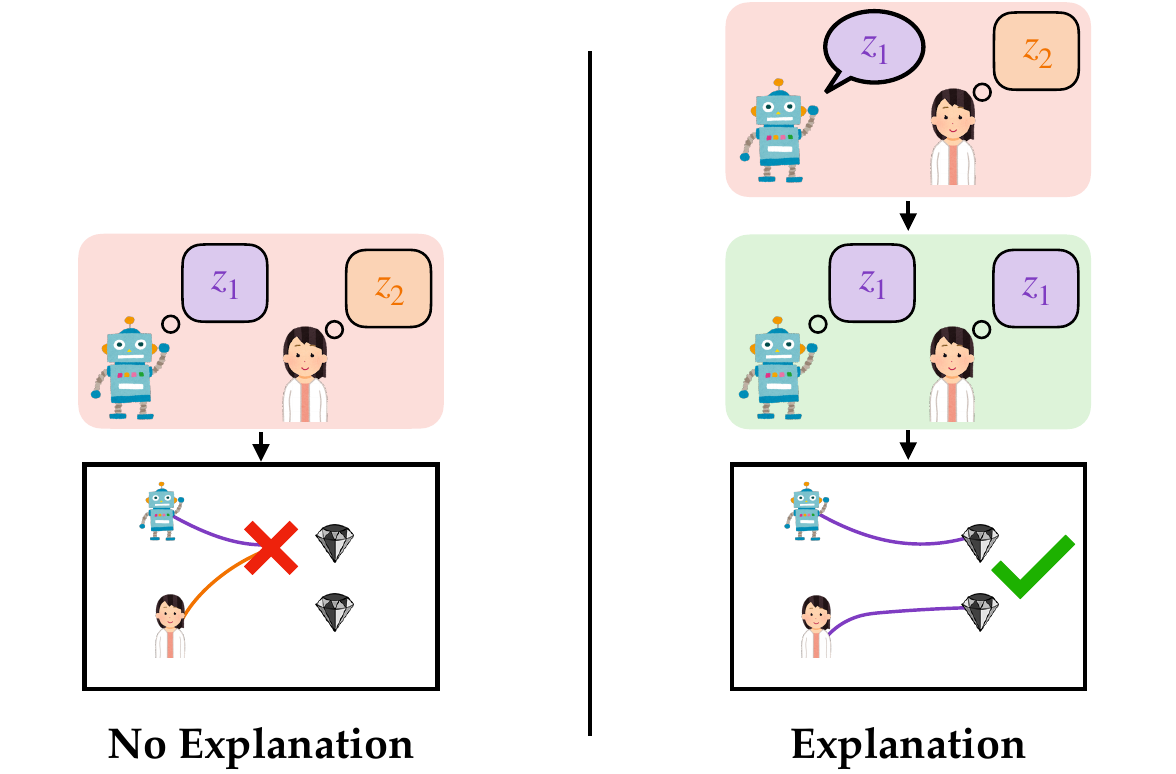}
    \caption{Depiction of how strategy explanations can solve the strategy alignment problem. The task is for each agent to grab one diamond without collisions. \textbf{Left:} The human and robot start with mismatched strategies, and ultimately collide \textbf{Right:} The human and robot start with mismatched strategies, but the robot explains strategy $z_1$, aligning with the human and resulting in a successful rollout.}
    \label{fig:front_figure}
    \vspace{-0.3in}
\end{figure}

In this work, we take a step towards enabling robots with the same capability: \textit{coordinating strategies} for \textit{collaborative tasks} with people (\figref{fig:front_figure}). As robots start being deployed alongside people in social, home, and manufacturing settings, it is increasingly important for robots to be able to collaborate fluently with and around people. This is particularly true when there are multiple potential strategies for collaboration, since we need to make sure all agents take the same strategy to complete the task efficiently.

We examine how collaborative fluency can be facilitated by a robot through a combination of proactive and reactive processes that communicate strategies for coordination \textit{a priori} but adapt to human strategies \textit{in situ}.
In this way, effective collaboration is not merely a reactive process where a robot purely adapts to the goals and behaviors of a human partner, but is a composite of reactive and proactive behaviors where the robot can leverage communication and suggestion to guide collaborations while remaining adaptive to the human in execution. Toward this goal, we examine how a robot can communicate particular strategies by distilling strategies into a few states.

The contributions of this work are:
\begin{enumerate}
    \item A general method for generating visual strategy explanations for human-AI collaboration in games with multiple Nash equilibria;
    \item A user study demonstrating that our method improves the ability of real users to collaborate efficiently with autonomous partners in a one-shot setting.
\end{enumerate}

\section{Related Work}

\noindent\textbf{Machine Teaching:} One class of approaches for generating explanations of a robot's behavior, called machine teaching \cite{zhu2018overview}, tries to generate a minimal set of examples that will allow the assumed learner model to understand how decisions were made. Originally applied to classification and regression tasks \cite{zhu2015machine, shinohara1991teachability}, such techniques have recently been used to generate trajectory demonstrations \cite{huang2019enabling, lee2021machine} for learners using inverse reinforcement learning \cite{ng2000algorithms}. Such approaches assume that a robot is either teaching a human how to perform a task or explaining how a robot would perform a task in isolation, but prior work lacks methods to generate multi-agent demonstrations for future collaborations.

\begin{figure*}
    \centering
    \includegraphics[width=0.9\textwidth]{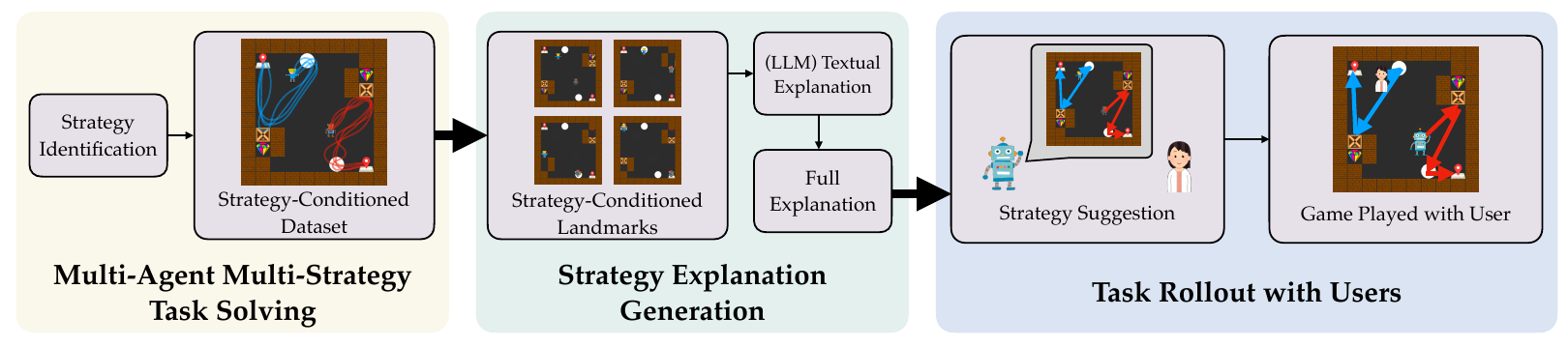}
    \caption{Pipeline for generating collaborative strategy explanations for a human. First, we identify strategies by clustering over the space of demonstrated examples. Then, we generate the set of  landmark states to summarize each strategy cluster and generate textual descriptions with an LLM. Finally, the robot shows the human user the generated explanation before collaboration.}
    \label{fig:generation_pipeline}
    \vspace{-0.2in}
\end{figure*}

\noindent\textbf{Policy Summarization:} A second class of explainability methods generates informative examples of a learned policy that may allow a human observer to understand its decisions, such as by finding states with high-entropy action distributions \cite{huang2018establishing}, by highlighting the ``importance'' of particular states \cite{guo2021edge} or by generating a reward map \cite{blakeman2022generating}. Other related work creates a graph of possible paths to a goal state and summarizes the set of necessary states to pass through \cite{sreedharan2020tldr}. We note that the literature has focused on summarizing \textit{single-agent} tasks, whereas our approach can be thought of as a multi-agent extention of policy summarization approaches for \textit{collaborative} tasks. 

\noindent\textbf{Human-Robot Collaboration in Multi-Strategy Tasks:} Much work has examined how robots can collaborate with human partners \cite{ajoudani2018progress}. Recent work allowed robots to replan online to react to the strategies of other agents \cite{fridovich2020efficient, cleac2019algames, wang2022co}, sometimes doing explicit inference \cite{peters2020inference} and even considering safety \cite{fisac2018probabilistically}. Other work has considered online information gathering of the human's strategy (or reward function) \cite{sadigh2016information, sadigh2016planning}, but these approaches rely on collision-avoidance behavior of humans to influence them. We note that these are largely \textit{reactive} approaches to collaborating with humans, but we are interested in enabling robots to proactively choose collaborative strategies with human partners. Other work has been done in the space of proactively influencing multi-agent interactions \cite{xie2021learning, wang2022influencing}, but are either not tested with real users or are too brittle to interact fluently with them.

\section{Multi-Agent Multi-Strategy Games}


\subsection{Problem Formulation}
\label{sec:problem_formulation}
We focus on a class of collaborative tasks where there may be multiple strategies (or equilibria) for completing the task. 
In particular, we consider $N-$player games represented by an HiP-MDP (Hidden Parameter Markov Decision Process) $\MM=\langle\SS, \mathcal{A}, \ZZ, \TT, \mathcal{R} \rangle$ where $\SS\subseteq\RR^{n}$ is the joint state space, $\mathcal{A}\subseteq\RR^{m}$ is the joint action space, $\ZZ\subseteq\RR^q$ is the hidden parameter space (or \textit{latent strategy} space) where $q\leq n$, $\TT$ is the transition dynamics, and $\mathcal{R} : \SS \times \AA \times \SS \rightarrow \RR^{N}$ is the union of all agents' reward functions. 
We assume that there are $N-1$ robot agents interacting with 1 human agent.

A joint policy $\pi: \SS \times \ZZ \rightarrow \AA$ defines the actions all agents should take in a particular state conditional on the joint strategy $z$: $a^t \sim \pi(s^t \mid z)$, 
where $a^t$ is the joint action of all agents and $s^t$ is the joint state of the system. The joint policy is the union of all agents' individual policies $\pi_i$ that dictate the individual actions $a_i^t$ of agent $i$ under their strategy $z_i$: $a^t=[a_1^t,...,a_N^t]^T = [\pi_1(s^t \mid z_1),...,\pi_N(s^t \mid z_N)]^T$.

\subsection{Objective}
\label{sec:objective}
Each agent has their own individual running reward function $r_i(\cdot)$. The agent's overall objective is to maximize their sum of expected rewards: $R_i = \sum_t r_i(s^t, a_i^t, s^{t+1})$.
The optimal solution to the HiP-MDP is a global Nash equilibrium, i.e. one that finds the policies $\pi^*_i$ and associated strategies $z^*_i$ where the sum of the agents' rewards are maximized so that no agent can change strategies to unilaterally improve their reward. Given an all agents' strategy vectors $\{z_1^*,\ldots,z_N^*\}$, we will know their rewards on the task $R_i(z_1^*,\ldots,z_N^*)$, so we can mathematically say (with a slight abuse of notation) that a set of strategies  is a Nash equilibrium if:
\begin{equation}
\begin{split}
    &R_i(z_1^*,\ldots,z_i^*,\ldots,z_N^*) \\
    &\geq R_i(z_1^*,\ldots,z_i,\ldots,z_N^*), \forall z_i\in \ZZ, \forall i\in [N].
\end{split}
\end{equation}

Solving for a global Nash equilibrium is generally intractable, so a joint policy $\pi^*$ for all agents could be obtained through reinforcement learning \cite{wang2022co, xie2021learning}, imitation learning \cite{le2017coordinated}, or by solving a dynamic game \cite{fridovich2020efficient, cleac2019algames}. These kinds of games in the real world often have multiple classes of (local) Nash equilibrium strategies.

\subsection{Multi-Agent Strategy Alignment}
We study the scenario where $N-1$ robots are collaborating with one human partner, which makes it difficult to roll out an optimal joint plan for all agents. The two main challenges here are that 1) humans may not be able to compute Nash equilibrium strategies, particularly in repeated games \cite{wright2010beyond, bornhorst2004people, bruttel2012infinity, gachter2004behavioral}, and 2) many scenarios of interest have multiple different equilibria, so the human needs to know which one the agents may have decided on (this has been called the \textit{strategy alignment} problem). 
All of the robot agents could coordinate with each other ahead of time, but they need some mechanism to coordinate with the human. 
Prior work uses online replanning to collaborate with humans \cite{fridovich2020efficient, peters2020inference}, but this turns the robot into a \textit{reactive} agent instead of a \textit{proactive} one. It may be useful to allow the human to lead the team's strategy, but we want robots to \textit{influence} the strategy that the team chooses. This can allow the human partner to choose strategies they may have otherwise been unfamiliar with. 
 

\section{Method: Strategy-Conditioned Landmarks}
\label{sec:strategy_explanations}
The full pipeline for our explanation generation method is illustrated in \figref{fig:generation_pipeline}. 
We note that many collaborative tasks contain discrete modes, even if the full space of strategies $\ZZ$ is continuous. We thus start by clustering $\ZZ$.

\subsection{Strategy Clustering}
\begin{figure}
    \centering
    \includegraphics[width=0.48\textwidth]{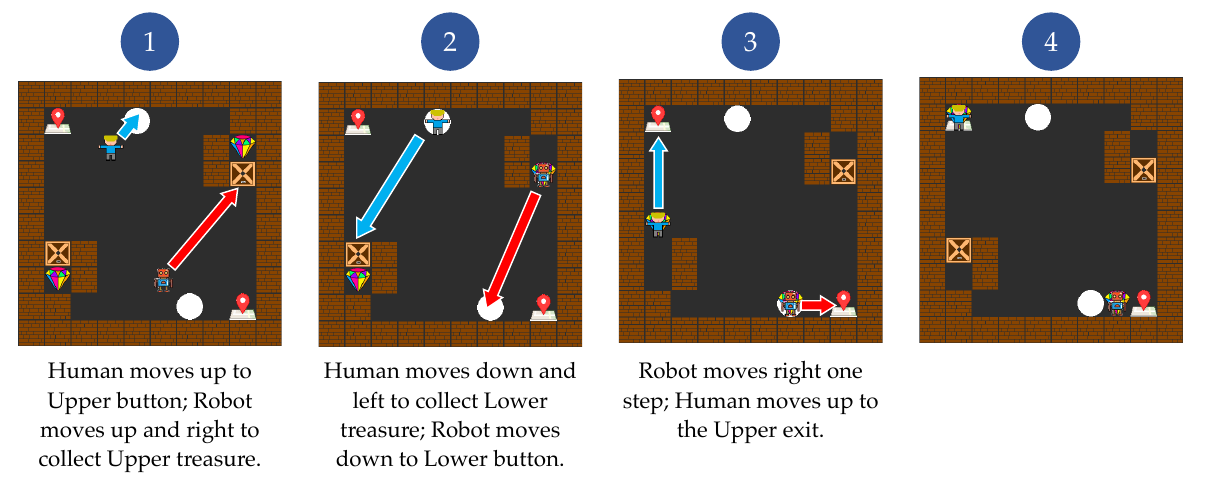}
    \caption{An explanation of one strategy generated on the collaborative maze task. Each image is a strategy landmark state computed from our method \sref{sec:strategy_explanations} and the text was generated from an LLM.}
    \label{fig:example_explanation}
    \vspace{-0.1in}
\end{figure}

\begin{figure}
    \centering
    \includegraphics[width=\columnwidth]{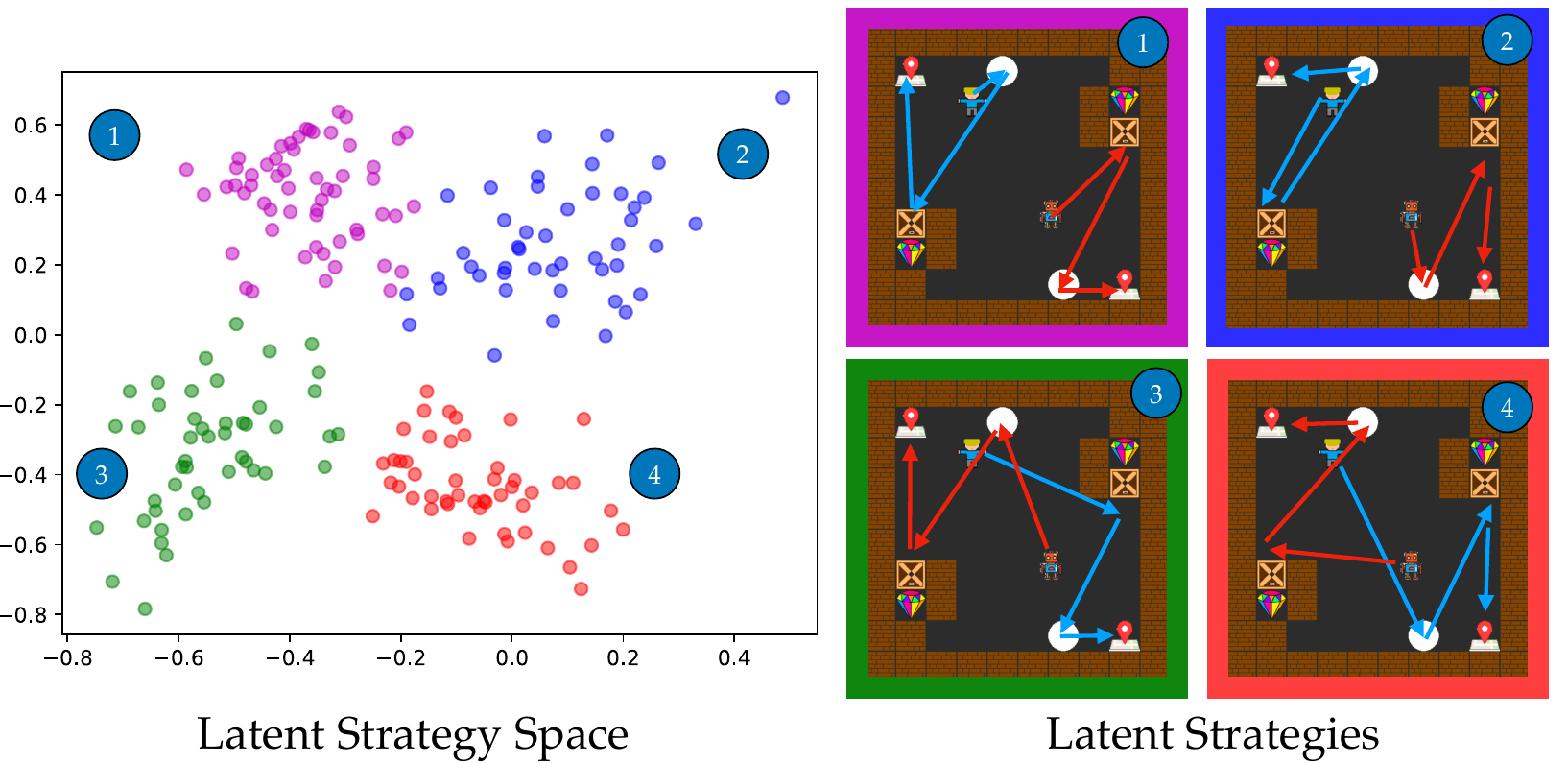}
    \caption{\textbf{Left:} Clustered latent strategy space for maze navigation. \textbf{Right:} Resulting strategies from each cluster center are a different ordering of the human and robot picking each jewel vs. standing on each button.}
    \label{fig:cogail_strategy_clusters}
    \vspace{-0.2in}
\end{figure}

To cluster the space of potential strategies, we assume access to a dataset $\DD$ containing trajectories $\xi_i$ of all $N$ agents performing the task $\DD=\{\xi_1,\ldots,\xi_l\}$ and their associated strategy vectors $\LL=\{z_1,\ldots,z_l\}$. Such a dataset may come from humans collaborating on the task or from rolling out a policy $\pi^*$ that solves the game multiple times (such as a differential game solver that outputs the Nash equilibrium policies for all agents). 

We can then cluster the strategy vectors $\LL$\footnote{In some scenarios, the latent strategy vectors may just be flattened trajectories, which works fine for our method.} using k-means clustering. We use silhouette analysis \cite{rousseeuw1987silhouettes} as a method for determining the number of strategy clusters. Resulting clusters of strategy vectors for the collaborative maze task (\sref{sec:maze_task}) are visualized in \figref{fig:social_nav_strategies}. Ultimately, this gives us a discrete set of strategy cluster centers $[z_1^*,...,z_k^*]$.

\vspace{-0.06in}
\subsection{Strategy Landmark State Generation}
\label{sec:landmark_states}
\vspace{-0.01in}
Prior work \cite{sreedharan2020tldr} has shown that for MDPs that can be written as stochastic shortest-path problems, one can compute a representative subset of the state space called \textit{landmark states} that every possible rollout of the policy must pass through in order to complete the task. We extend this idea to the HiP-MDPs we consider in this work by utilizing the previously discussed clustering of the latent space to generate \textit{strategy-conditioned landmark states}.

\begin{definition}[Strategy-Conditioned Policy Landmark State]
Given an MDP $\MM$ and a strategy cluster center $z_i^*$, and a dataset of trajectories $\DD=\{\xi_1,\ldots,\xi_l\}$, a strategy-conditioned policy landmark state $l$ (or simply \textit{strategy landmark}) is a state (or cluster of states) that all trajectories $\xi_j$ in the set of trajectories corresponding to the strategy cluster $z_i^*$ (called $\DD_{z_i^*}$) pass through. 
\end{definition}

To find the set of strategy landmarks, we first discretize the state space then cluster the discretized trajectories from each strategy $\DD_{z_i^*}$. This results in a set of landmark states for each strategy cluster: $S_{land}^i = \{l_i^1,\ldots,l_i^{k_i}\}$ where $k_i$ a hyperparameter that determines the number of clusters and landmark states. One set of landmark states for the collaborative maze task (\sref{sec:maze_task}) can be seen in \figref{fig:example_explanation}.

\subsection{Textual Explanation Generation}
\label{sec:textual_explanation}
During initial pilot studies, many users expressed that some amount of textual explanation would be useful for better understanding the strategies. We first wrote out the rules of each game and translated the strategy landmarks generated from \sref{sec:landmark_states} to natural language. We then asked a large language model (ChatGPT) to generate descriptions for what happened between pairs of strategy landmarks $(l_i^k,l_i^{k+1})$\footnote{In the future, it may be more scalable to use a vision-language model (VLM) to generate explanations so as to avoid manual task translation.}. 
Text for one strategy is shown in \figref{fig:example_explanation}.

\subsection{Baseline: Video Explanation}
\label{sec:vid_explanation}
Since existing policy summarization methods do not explicitly deal with collaborative tasks, we construct an intuitive baseline of showing policy rollouts directly. For each strategy cluster $\mathcal{D}_{z_i^*}$, the video explanation baseline is a video of the trajectory corresponding to the strategy cluster center $z_i^*$. For example, videos for the collaborative maze task show rollouts of the trajectories depicted in \figref{fig:cogail_strategy_clusters} \textbf{right}.

\subsection{Modification: Landmark Video Explanation}
\label{sec:vid_landmark}
To make the most fair comparison, we additionally augment our method with video explanations. Specifically, we freeze the video explanation from strategy cluster $z_i^*$ at each landmark state $l_i^j$ for 10 seconds and display explanation text.


\begin{figure*}
    \centering
    \includegraphics[width=0.9\textwidth]{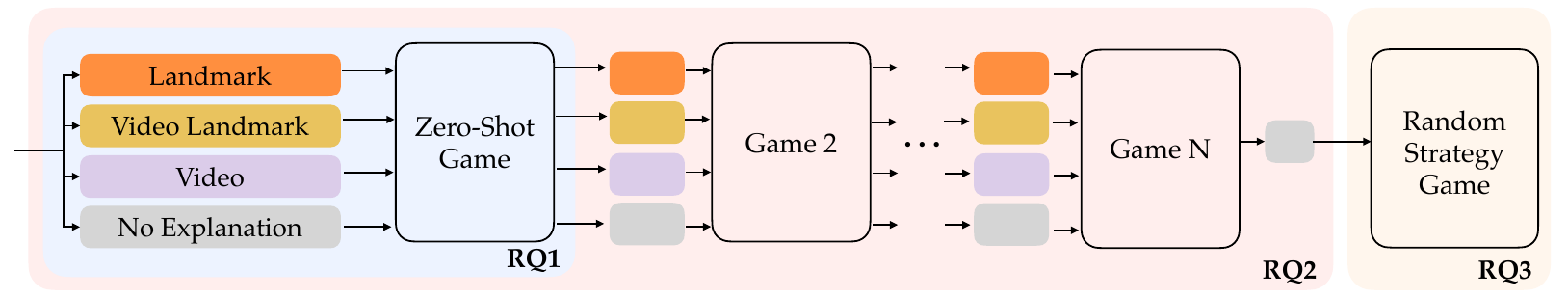}
    \caption{A the three-part user study for testing our approach. First, users are assigned an explanation type: \textit{landmarks}, \textit{landmark videos}, \textit{videos}, or \textit{no explanation}. For each of two environments, users play $N$ games with one explanation between each (or none in the \textit{no explanation} case) and finally play one game with no prior communication with an agent that takes a random strategy.}
    \label{fig:study_flow}
    \vspace{-0.25in}
\end{figure*}

\section{Collaborative Maze Task}
\label{sec:maze_task}
For our first experimental domain, we use a 2D maze task built by previous work \cite{wang2022co}. In this task, two agents need to collaborate to retrieve one treasure each and then take them to the exits. The agents need to coordinate their actions because the treasures are blocked by doors that can only be opened while one agent is standing on a nearby button. The task is shown in \figref{fig:cogail_strategy_clusters}. The agents are rewarded when both agents reach different goals holding treasures.

\subsection{Co-GAIL}
Prior work \cite{wang2022co} collected a dataset $\DD$ of pairs of humans completing this task and used this data to train a policy (via their algorithm called Co-GAIL) that can output actions for the different strategies that humans exhibited. 
We utilize the latent strategy space learned by Co-GAIL to generate strategy clusters for strategy landmark generation (\sref{sec:strategy_explanations}).

In an initial pilot study, we directly used the Co-GAIL model with real users. However, we found it was often too brittle to always finish the task (in strategy 2 in \figref{fig:cogail_strategy_clusters}, the robot completed the game with only 25\% of participants). In the original paper \cite{wang2022co}, the authors report 85-100\% success rates with people, but only test with 4 human users repeatedly interacting with the system. Users likely became familiar with how to best play the game with this particular agent. 

\subsection{Robustly Acting with Human Collaborators}
\label{sec:robustly_acting}
We created one robot planner for each strategy cluster that moves between the landmark states. Since the strategy clusters were still found using the latent codes output by the model trained with Co-GAIL, this approach allows us to take advantage of learning-based techniques while still robustly and reliably completing the task with new users. The planner $\mathcal{P}$ takes in a pair of sequential landmark states for a particular strategy and outputs the robot's action: $a_R^t = \mathcal{P}(l_i^j, l_i^{j+1}, t)$.
Once the human and robot reach the strategy landmark state $l_i^{j+1}$, the planner moves to planning for $l_i^{j+2}$.

In this setting, we still want to measure how much the robot's explanations are able to influence the person by measuring their adherence to the planned strategy. However, due to the nature of the task, the human and robot necessarily need to coordinate their actions in order to complete the task. We thus employ a basic Bayesian inference method over the first landmark state of each strategy to allow the robot to correctly respond to the actual chosen human strategy to complete the task:
\begin{equation}
    b_R^{t+1}(l^1) = \frac{p(a_H^t\mid s_H^t; l^1) b_R^t(l^1)}{\sum_{\hat{l}^1} p(a_H^t\mid s_H^t; \hat{l}^1) b_R^t(\hat{l}^1)}.
\end{equation}
where the assumed reward for the human in the likelihood function $p(a_H^t\mid s_H^t; l^1)$ is the negative distance traveled. 

\section{Social Navigation Task}
\label{sec:social_nav_task}
\begin{figure}
    \centering
    \includegraphics[width=\columnwidth]{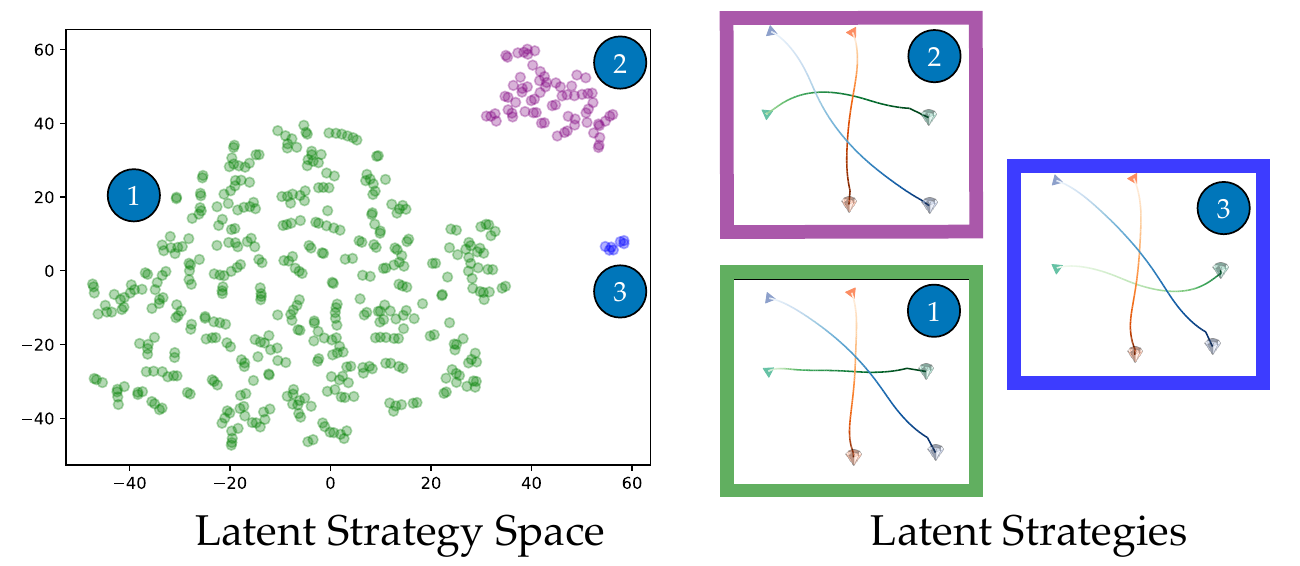}
    \caption{\textbf{Left:} Clustered latent strategy space for social navigation task. \textbf{Right:} Resulting strategies from each cluster center. The agents start at the locations denoted by the triangles and the goal is to reach the diamond of the corresponding color without colliding.}
    \label{fig:social_nav_strategies}
    \vspace{-0.2in}
\end{figure}

\subsection{Problem Setup}
\label{sec:navigation_problem}
Our second experimental domain is a 2D social navigation task where $N$ dynamic agents have different goals and need to reach them while avoiding collisions. Since we're dealing with dynamical systems, it is natural to represent them as ODEs in continuous time. In practice, the systems are integrated with a discrete time interval of $t_s=0.1s$ using a Runge-Kutta method (RK4). This allows the dynamical system to be in the same form as the HiP-MDP outlined in \sref{sec:problem_formulation}. All $N=3$ agents have continuous unicycle dynamics $\dot s_i = f(s_i) + g(s_i)a_i$.
The state $s_i$ consists of the $(x,y)$ position $(p_x,p_y)$, the velocity $v$ and the steering angle $\psi$, and the control action is the acceleration and steering angle change $a_i=[\dot v, \dot\psi]^T$. The MDP state is $s^t=[s_1^{t,T}, s_2^{t,T}, s_3^{t,T}]^T$. The rewards are as detailed in \cite{fridovich2020efficient}.


\subsection{ILQGames}
To solve the game posed in \sref{sec:navigation_problem}, we use the LQ-game solver proposed in \cite{fridovich2020efficient} that solves for local Nash equilibria. Specifically, we use the implementation in Julia---iLQGames.jl \cite{peters2020ilqgames}. 
Given an initial condition $s^0$, agent dynamics, and agent costs, the solver returns a trajectory $\xi$ and control policy (strategy) $z$ for all agents. We perturb the initial condition $s^0$ 500 times and run the solver on each initial condition to get a diverse dataset of multi-agent interactions $\DD=\{\xi_1,\ldots,\xi_{500}\}$, $\LL=\{z_1,\ldots,z_{500}\}$. A silhouette analysis tells us there should be 3 clusters in the strategy space. We can see the resulting three strategy clusters in \figref{fig:social_nav_strategies}. The plotted latent strategy space shows a 2D t-SNE projection of the full latent strategies.

Equipped with a dataset $\DD$ of multi-agent trajectories and an associated set of cluster centers $\{z_1^*,\ldots,z_3^*\}$, we generate explanations for each strategy in this task as detailed in \sref{sec:strategy_explanations}. Here, it is easy to directly compute whether the human adhered to the planned strategy by checking which existing strategy cluster the rolled-out trajectory belongs to (since completion of the game does not depend on coordinated actions). We thus have the robot agents roll out their policies without inferring the human's strategy online.

\section{User Studies}



\noindent\textbf{Research Questions.} We conducted an in-person user study to answer the following research questions: (\textbf{RQ1}) How does the explanation type affect human performance and collaborative fluency in a zero-shot coordination task? (\textbf{RQ2}) Does the explanation type influence whether people adhere to the robot’s suggested strategy? (\textbf{RQ3}) Does robot-guided strategy exploration improve humans’ abilities to collaborate in subsequent games \textit{without} prior strategy coordination?

\begin{figure}
    \centering
    \includegraphics[width=\columnwidth]{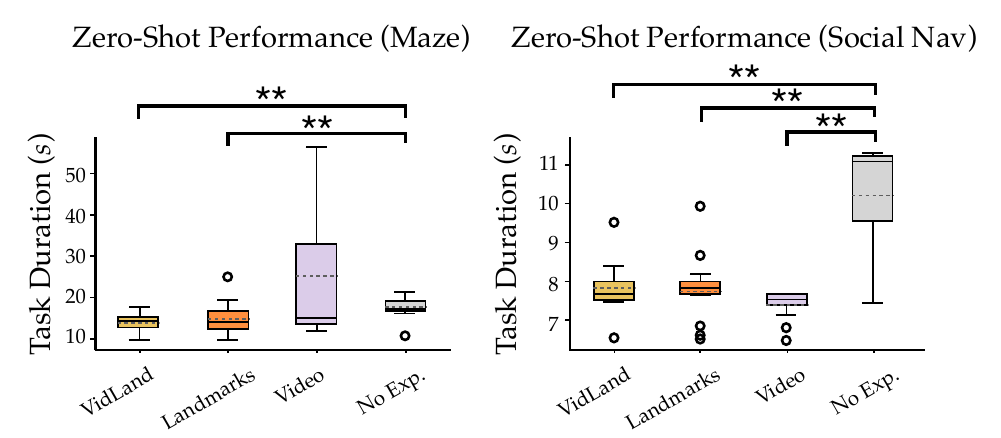}
    \caption{Task duration on study. Gray dotted: mean, black line: median.}
    \label{fig:task_duration}
    \vspace{-0.2in}
\end{figure}

\begin{figure}
    \centering
    \includegraphics[width=\columnwidth]{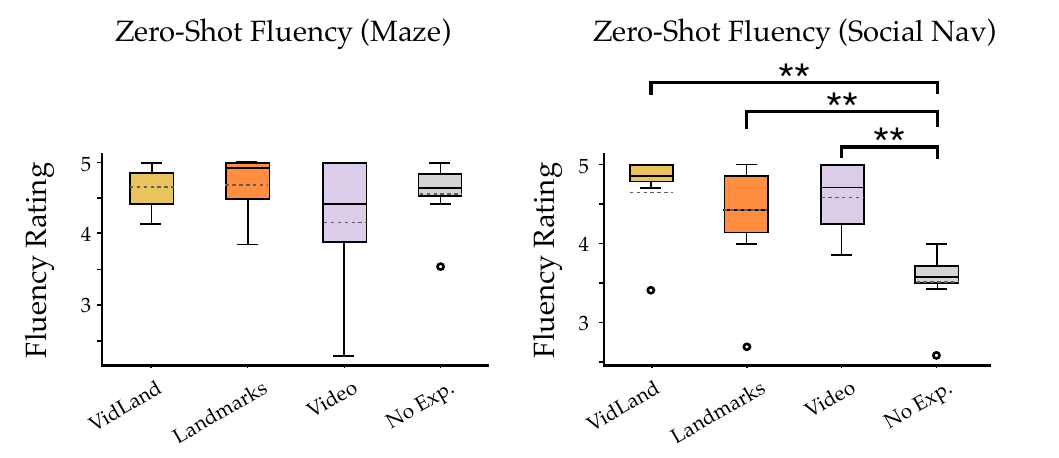}
    \caption{Subjective fluency ratings}
    \label{fig:subjective_fluency}
    \vspace{-0.2in}
\end{figure}

\noindent\textbf{Experimental Setup:} 
We design a three-part study (\figref{fig:study_flow}) to answer all three research questions. All participants are given the same set of instructions for how to play each game before being guided through a series of collaborative tasks with the robot. Participants play the games using the joystick on a video game controller (Nintendo Switch Pro Controller). 

In part 1 (P1), participants are given an explanation of one strategy by the robot (or no explanation in the baseline condition), then play one game with the robot. In part 2 (P2), the robot guides the participant through the rest of the strategies in a randomized order. Finally, in part 3 (P3), once the guided strategy exploration is complete, the participant plays the game with a new robot partner who executes a random strategy without explicitly communicating it. Participants experience this three-part study sequence for both the Maze (\sref{sec:maze_task}) and Social Nav (\sref{sec:social_nav_task}) tasks.

\noindent\textbf{Independent Variables:} We vary the explanation type with 4 levels: \textit{landmark images} (\sref{sec:strategy_explanations}), \textit{landmark videos} (\sref{sec:vid_landmark}), \textit{videos} (\sref{sec:vid_explanation}), and \textit{no explanation}. 

\noindent\textbf{Objective Measures:}
\textit{Task performance} (relevant to P1 and P3) measures how long it takes for the human-robot team to complete the task. \textit{Strategy adherence} (P2) is measured by the number of times the human chose to follow the robot's suggested strategy. The no-explanation condition is removed from this analysis because the robot does not provide strategy guidance. \textit{Strategy exploration} (P2) is measured by the number of unique strategies played by the team in each of the games prior to the final round with the unseen partner. 

\noindent\textbf{Subjective Measures:} \textit{Subjective fluency} (P1) is a measure of how well the human thinks the team   coordinated their actions in the task, and is self-reported by the participant through seven Likert-scale questions drawn from \cite{hoffman2019evaluating} (Cronbach's $\alpha=0.904$ for Maze, and $\alpha=0.913$ for Social Nav).

\noindent\textbf{Hypotheses}. We intuit that  landmark-based explanations will improve participants' understanding of the strategies, compared to video-only explanations which may contain noise. We expect improved task performance, adherence to the robot's suggestion, and coordination with an unseen partner. \textbf{H1}: Participants guided by landmark images and landmark videos will complete the zero-shot coordination task (in both Maze and Social Nav) faster than the other conditions (P1).  \textbf{H2}: Participants guided by landmark images and landmark videos will adhere more to the robot's guidance than the other conditions (P2). \textbf{H3}: Participants guided by explanations will execute more unique strategies than those not shown explanations (P2). \textbf{H4}: Participants guided by landmark images and landmark videos will be able to adapt to an unseen partner's strategy faster than the other conditions in both environments (P3). 

\noindent\textbf{Participants:} We recruited 49 in-person participants from the campus community and CMU's Center for Behavioral and Decision Research participant pool, of which 49.0\% self-identified as female, 44.9\% as male, and 6.1\% as non-binary. Participant ages ranged from 18 to 70 (M = 25.7, SD = 9.4). We removed participant data in each task if they experienced technical errors, giving us 43 participants, divided as: in the maze task, 14  in  \textit{landmark images} condition, 11 in \textit{landmark videos}, 10 in \textit{videos}, and 8 in \textit{no explanation}; in the social navigation task, 13  in \textit{landmark images}, 11 in \textit{landmark videos}, 12 in \textit{videos}, and 7 in \textit{no explanation}.






\noindent\textbf{Collaborative Maze Results:}
For each measure, we removed outliers using the interquartile range method. We used one-way ANOVAS for each measure, and used post-hoc tests with Bonferroni correction to examine main effects. In all figures, (*) means $p<0.05$ and (**) means $p<0.01$.

\textbf{H1} was partially supported. We found a significant difference between explanation types on zero-shot coordination performance (\figref{fig:task_duration}), measured by the time taken to complete the game ($F(3,37)=4.11, p=0.013, \eta^2 = 0.25$). Post-hoc analysis showed that participants guided by landmark images ($p_{bonf}<0.01$) and landmark videos ($p_{bonf}<0.01$) completed the zero-shot coordination task significantly faster than those not shown any explanation, while those guided by videos showed no significant improvement over no explanation. 
There was no significant difference between conditions in subjective collaborative fluency (\figref{fig:subjective_fluency}) in the zero-shot interaction ($F(3,38)=2.06, p=0.122, \eta^2 = 0.14$). However, we observe that the variance in subjective fluency and task duration was higher for video explanations. 

\begin{figure}
    \centering
    \includegraphics[width=\columnwidth]{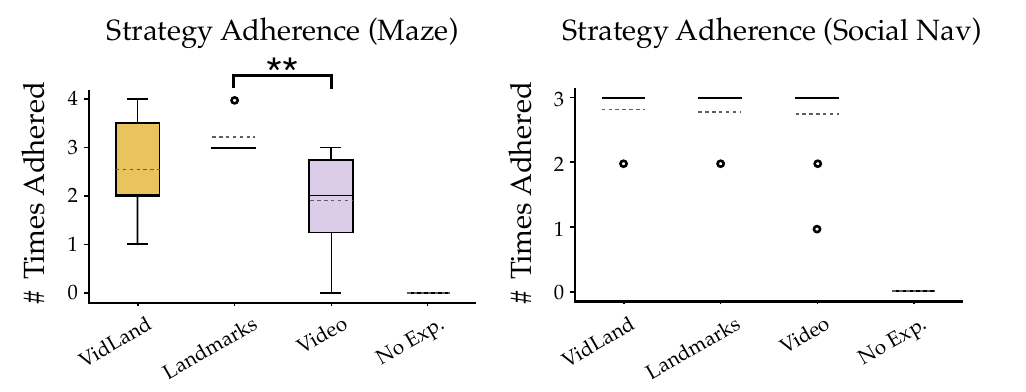}
    \vspace{-0.2in}
    \caption{Adherence to explained strategy by robot}
    \label{fig:strategy_adherence}
    \vspace{-0.1in}
\end{figure}

\textbf{H2} was supported. Explanation type significantly affected adherence to the robot's suggested strategies (\figref{fig:strategy_adherence}, $F(2,32)=6.795, p<0.01, \eta^2 = 0.30$). Participants followed the robot's guidance significantly more often when presented strategies in landmark image form than in video form ($p_{bonf}<0.01$). 

\begin{figure}
    \centering
    \includegraphics[width=\columnwidth]{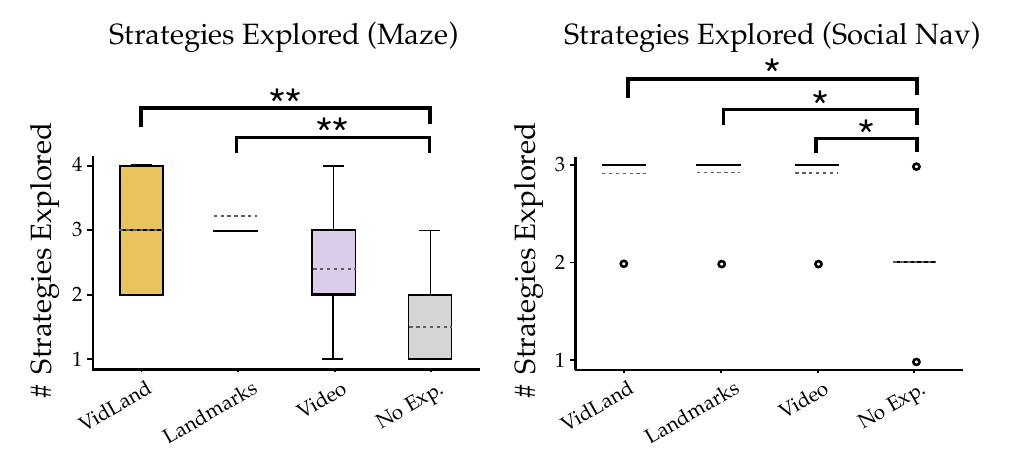}
    \vspace{-0.2in}
    \caption{Number of strategies explored in repeated rounds.}
    \label{fig:num_strat_explored}
    \vspace{-0.25in}
\end{figure}

\textbf{H3} was supported. Explanation type significantly affected how many unique strategies were actually performed by the team (\figref{fig:num_strat_explored}, $F(3,39)=7.367, p<0.01, \eta^2 = 0.36$). Landmark videos ($p_{bonf}<0.01$) and landmark images ($p_{bonf}<0.01$) both had significantly more unique strategies compared to the no explanation condition. Participants without any robot guidance tended to not explore strategies; they repeatedly stuck with whichever strategy was performed first for games. Video demonstrations did not significantly influence people to explore the strategies in the games played. Although there was no significant difference between video and landmark video results, the means for both zero-shot task performance and number of strategies explored are higher for landmark videos and landmark images than for videos alone. This suggests that landmarks enhance participant understanding more effectively than videos. 

\begin{figure}[t]
    \centering
    \includegraphics[width=\columnwidth]{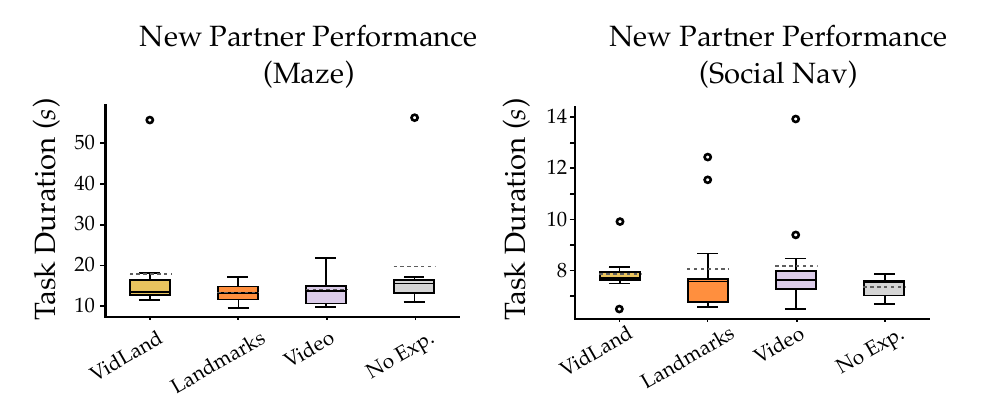}
    \vspace{-0.2in}
    \caption{Task duration with new (random) partner.}
    \label{fig:new_partner_duration}
    \vspace{-0.25in}
\end{figure}

\textbf{H4} was not supported. We did not find a significant difference in task performance  with an unseen (random strategy) partner by explanation type (\figref{fig:new_partner_duration}, $F(3,37)=0.348, p=0.79, \eta^2 = 0.03$). Since the starting state and environment do not change, practicing the task four times likely enabled participants to adapt  to the random strategy partner.

\noindent\textbf{Social Navigation Results:}

\textbf{H1} was partially supported in the social navigation task. Explanation type significantly affected the duration of the first game  (\figref{fig:task_duration}, $F(3,30)=20.69, p< 0.01, \eta^2 = 0.67$). Participants who were provided video ($p_{bonf}=0.04$), landmark videos ($p_{bonf}=0.03$), and landmark images ($p_{bonf}=0.04$) all performed better than those not offered any strategy explanation. 
In other words, participants guided by explanations of any form (video, landmark videos, and landmark images) completed the zero-shot coordination task significantly faster than those not shown any explanation.
Similarly, explanation type had an effect on  subjective collaborative fluency ratings (\figref{fig:subjective_fluency}, $F(3,35)=7.07, p< 0.01, \eta^2 = 0.38$).
The participants who were provided video ($p_{bonf}<0.01$), landmark videos ($p_{bonf}<0.01$), and landmark images ($p_{bonf}<0.01$) all found the collaboration more fluent than those not offered any strategy explanation. 

\textbf{H2} was not supported. There was no significant difference in adherence to the robot's suggested strategies by explanation type (\figref{fig:strategy_adherence}, $F(2,33)=0.056, p=0.94, \eta^2 = 0.003$). 

\textbf{H3} was supported. Explanation type had a significant effect on how many unique strategies were actually performed by the team (\figref{fig:num_strat_explored}, $F(3,39)=13.48, p<0.01, \eta^2 = 0.51$). Landmark videos ($p_{bonf}=0.03$), landmark images ($p_{bonf}=0.03$), and videos ($p_{bonf}=0.03$) all significantly increased the number of unique strategies the team executed compared to the no explanation condition. 

\textbf{H4} was not supported. We did not find a significant difference in task performance with the unseen (random strategy) partner based on explanation type (\figref{fig:new_partner_duration}, $F(3,33)=1.30, p=0.29, \eta^2 = 0.11$). 

In the social navigation task, all explanations improved subjective fluency, task performance, and strategy exploration. Unlike in the maze task, each agent must only avoid---but does not rely on---the others. Landmarks and videos may provide similar benefits in such a simple task. 










\section{Conclusion and Future Work}
We proposed a method for strategy alignment in multi-agent tasks when there are multiple discrete modes of collaboration. Specifically, we defined a framework for generating visual strategy explanations via strategy-conditioned landmark states. We evaluated the proposed method on two experimental domains where a robot (or multiple robots) used strategy explanations to induce efficient collaborations with a human partner. We ultimately found promising results; users tended to perform better in their first collaboration with a robot when presented with a strategy explanation, particularly one using landmarks. Users also explored the space of possible strategies more when guided by the robot's explanations to do so. 

Though we did not find that explanations prepared users for future collaborations, it suggests that in more complex tasks, robots may be able to communicate collaborative strategies efficiently. 
Researchers may consider explanation generation using home simulators \cite{puig2018virtualhome, puig2023habitat}. One challenge when scaling up may be determining appropriate strategies to explain when the set of discrete strategy modes is large. Additionally, future work may try extending to physical human-robot collaborations, where strategy alignment has effects on the \textit{safety} of the interaction. 

\section*{Acknowledgments}
This material is based upon work supported by the National Science Foundation Graduate Research Fellowship under Grant No. DGE1745016 and DGE2140739 and additionally under Grant No. 2144489. Any opinions, findings, and conclusions or recommendations expressed in this material are those of the authors and do not necessarily reflect the views of the National Science Foundation.

\bibliographystyle{IEEEtran}
\bibliography{references}


\appendices
\section{Strategy-Critical States}
Previous work has looked at generating explanations for a human of when an autonomous agent thinks it's very important to take a particular action to not fail at the task as a series of states, called \textit{critical states}. Given a stochastic policy $\pi$, the original paper searches the set of critical states $\mathcal{C}_{\pi}$ where the agent prefers a small set of actions over all others:
\begin{equation}
    \mathcal{C}_{\pi} = \{s \mid \mathcal{H}(\pi(\cdot\mid s)) < t\}
\end{equation}
where $\mathcal{H}(\pi(\cdot\mid s))$ is the entropy of the policy's action distribution at state $s$ and $t\in\mathbb{R}$ is a threshold for criticality. 

We modify their approach to try and generate explanations for when the agent thinks it's very important to take particular actions to end up taking particular strategies. In particular, we consider finding the set of \textit{strtategy-critical states} $\SC_{\pi}$ where the agent prefers taking a small set of \textit{different actions} for \textit{each strategy}:
\begin{equation}
    \SC_{\pi} = \{ s \mid \frac{1}{k^2} \sum_{z_i^*,z_j^*} D_{KL}(\pi(\cdot | s, z_i)) \parallel \pi(\cdot | s, z_j) > t\}.
\end{equation}
where $D_{KL}$ is the KL-divergence. This method will search for states where the average KL-divergence between the action distributions for the policy under each potential strategy cluster are sufficiently large. 

Using this approach, we find that strategy-critical states are branching points between strategies, i.e. the last possible common state between different strategies. Most strategy-critical states show the two agents just before committing to a particular strategy. We use these strategy-critical states to determine the initial strategy landmark when generating the full explanation.

\section{Textual Explanation Generation}
\label{sec:text_explanation}
\textit{Game Description Prompt: }

A two-player game in a 10 by 10 grid. Row 0, column 0 is the top left corner. This is the starting state of the game.
\begin{itemize}
    \item H represents Player 1 at row 3, column 4
    \item R represents Player 2 at row 7, column 6
    \item J1 represents the jewel at row 3, column 9
    \item J2 represents the jewel at row 7, column 2
    \item B1 represents the button at row 2, column 5
    \item B2 represents the button at row 9, column 7
    \item E1 represents the exit at row 2, column 2
    \item E2 represents the exit at row 9, column 9
    \item D1 represents the door at row 4, column 9
    \item D2 represents the door at row 6, column 2
\end{itemize}

Here are the rules of the game. There is a jewel at row 3, column 9, but it is blocked by a door at row 4, column 9. There is a second jewel at row 7, column 2, but it's blocked by a door at row 6, column 2. In order to open the door at row 4, column 9, one of the two players must stand at the location of a button, which is located at row 2, column 5. When one of the players stands on the button, the door will open, and the other player can collect the jewel. In order to open the door at row 6, column 2, one of the two players must stand at the location of a second button, which is located at row 9, column 7. Additionally, once a player has collected a jewel, that player cannot collect the other jewel. The player who has already collected a jewel must help the other player to collect the other jewel. Once both jewels have been collected, one player needs to move to the location of an exit, located at row 2, column 2. And the other player needs to move to the location of a second exit, located at row 9, column 9. When both players are located at each of the exits simultaneously, the game ends.

\textit{Landmark Description Prompt: }
State B is the following. In State B, Player R is located at row 8, column 6. Player H is located at row 3, column 4. Next we will describe a different state, State C. In State C, Player H is at the location of the button at row 2, column 5. Player R is at row 4, column 9, and is holding a jewel. Describe succinctly (<15 words) and intuitively in present tense what happened to get from State B to State C.

\textit{ChatGPT Response: }
H moves up to Upper button; R moves up and right to collect Upper jewel.

\end{document}